\documentclass[]{elsarticle}

\usepackage{bbm,amssymb,amsmath,amsfonts,amsthm}
\usepackage{algorithm}
\usepackage{algpseudocode}
\usepackage{ifxetex,ifluatex}

\ifnum 0\ifxetex 1\fi\ifluatex 1\fi=0 % if pdftex
\usepackage[T1]{fontenc}
\usepackage[utf8]{inputenc}
\else % if luatex or xelatex
  \ifxetex
    \usepackage{mathspec}
    \usepackage{xltxtra,xunicode}
  \else
    \usepackage{fontspec}
  \fi
  \defaultfontfeatures{Mapping=tex-text,Scale=MatchLowercase}
  
\fi

\providecommand{\tightlist}{%
  \setlength{\itemsep}{0pt}\setlength{\parskip}{0pt}}

\IfFileExists{upquote.sty}{\usepackage{upquote}}{}
\IfFileExists{microtype.sty}{%
\usepackage{microtype}
\UseMicrotypeSet[protrusion]{basicmath} % disable protrusion for tt fonts
}{}

\usepackage{natbib}
\usepackage{graphicx}
\makeatletter
\def\maxwidth{\ifdim\Gin@nat@width>\linewidth\linewidth\else\Gin@nat@width\fi}
\def\maxheight{\ifdim\Gin@nat@height>\textheight\textheight\else\Gin@nat@height\fi}
\makeatother
\setkeys{Gin}{width=\maxwidth,height=\maxheight,keepaspectratio}
\ifxetex
  \usepackage[setpagesize=false, % page size defined by xetex
              unicode=false, % unicode breaks when used with xetex
              xetex]{hyperref}
\else
  \usepackage[unicode=true]{hyperref}
\fi
\hypersetup{breaklinks=true,
            bookmarks=true,
            pdftitle={Decomposing the Deep: Finding Class Specific Filters in Deep CNNs (v1)},
            colorlinks=true,
            citecolor=blue,
            urlcolor=blue,
            linkcolor=magenta,
            pdfborder={0 0 0}}
\begin{document}

%% Front matter
\begin{frontmatter}
\title{Decomposing the Deep: Finding Class Specific Filters in Deep CNNs}

%% affiliations is defined as a list of affiliation strings and inserted as is
%% For now including full string in affiliation
%% {organization, addresline (can be multiple), city, postcode, state, country}
%% pandoc template cannot zip so it's difficult to define it programmatically
\affiliation[uoh]{
  department={School of Computer and Information Sciences},
  organization={Univeristy of Hyderabad},
  addressline={Prof C R Rao Road, Gachibowli},
  city={Hyderabad},
  postcode={500019},
  state={Telangana},
  country={India}
}

%% Authors and affiliations are combined
%% Each author in [aff]{author}\ead{mail@whatever} format
\author[uoh]{Akshay Badola}
\ead{badola@uohyd.ac.in}
%% Each author in [aff]{author}\ead{mail@whatever} format
\author[uoh]{Cherian Roy}
\ead{19mcmi10@uohyd.ac.in}
%% Each author in [aff]{author}\ead{mail@whatever} format
\author[uoh]{Vineet Padmanabhan}
\ead{vineetnair@uohyd.ac.in}
%% Each author in [aff]{author}\ead{mail@whatever} format
\author[uoh]{Rajendra Prasad Lal}
\ead{rajendraprasd@uohyd.ac.in}

%% Additional Includes

%% Research Highlights

%% Keywords

%% Abstract
\begin{abstract}
Interpretability of Deep Neural Networks has become a major area of
exploration. Although these networks have achieved state of the art
results in many tasks, it is extremely difficult to interpret and
explain their decisions. In this work we analyze the final and
penultimate layers of Deep Convolutional Networks and provide an
efficient method for identifying subsets of features that contribute
most towards the network's decision for a class. We demonstrate that the
number of such features per class is much lower in comparison to the
dimension of the final layer and therefore the decision surface of Deep
CNNs lies on a low dimensional manifold and is proportional to the
network depth. Our methods allow to decompose the final layer into
separate subspaces which is far more interpretable and has a lower
computational cost as compared to the final layer of the full network.
\end{abstract}

\end{frontmatter}

%% Includes before main content

%% Main body
\hypertarget{introduction}{%
\section{Introduction}\label{introduction}}

Deep Neural Networks have been a paradigm shift in machine learning but
they have remained primarily black boxes. Large networks can contain
millions \cite{szegedy2015going}, \cite{he2016deep} to billions
\cite{radford2019language}, \cite{brown2020language} of parameters.
Such models are highly overparamterized and in fact, overparametrization
is essential to their generalization ability \cite{choromaska2015loss},
\cite{du2019gradient}, \cite{du2019gradient_a}. In such cases it
becomes extremely difficult to attribute the prediction of any network
to its parameters which leads to poor understanding and interpretability
of the network. The lack of interpretability of these models makes it
very difficult for humans to trust them in critical situations, like
medical diagnosis or autonomous vehicles. It also becomes very hard to
diagnose and correct the models themselves.

Final layers\footnote{When we say final layer(s) we mean the layers
  closer to the classifying layer. Following previous works we also
  mention them as \emph{top} layer(s) and the layers closer to input as
  \emph{bottom} layer(s).} of a CNN hold particular interest for
interpretability as detectors that appear there are more aligned with
semantic concepts. The final layer of the network projects the features
via softmax onto the decision simplex and its properties are critical to
the final label predicted by the network.

While there has been recent work in interpreting and disentangling
neural network features; in the current literature we have not come
across an efficient method to identify and rank the features in the
final and penultimate layers and consequently decompose the layer. In
this work we focus on these final and penultimate layers of some popular
CNN architectures, namely Resnet \cite{he2016deep}, Densenet
\cite{huang2017densely}, Efficientnets \cite{tan2019efficientnet}. We
illustrate some properties of those layers and consequent features and
provide an efficient method to decompose the final layer which results
in interpretability via disentanglement and reduces the computational
complexity of the final layer. We also provide the implementation for
decomposing the final layers\footnote{We have not given the code link to
  preserve the authors' anonymity for the review process and will
  include it post review.}.

\hypertarget{related-work}{%
\section{Related Work}\label{related-work}}

Zhang et. al. \cite{zhang2020survey_a} define interpretability as an
``ability to provide explanations in understandable terms to a human''.
We however, leave it to humans to develop the terms necessary for
communication and instead define interpretability as \emph{a
representation of a model amenable to inspection, adaptation and
attribution}.

An interpretable model, and specifically an interpretable Deep Neural
model in our case, should be easy to inspect, easy to adapt for and easy
to attribute. In particular it should facilitate:

\begin{enumerate}
\def\labelenumi{\arabic{enumi}.}
\tightlist
\item
  Input inspection and attribution: An interpretable model should allow
  correspondence between parts of an input to the labels. The
  correspondence may be simplified for the purpose. For example in an
  image, parts of an object which contribute more towards the
  classification could be filtered for inspection and attribution.
\item
  Feature attribution: An interpretable model should allow
  correspondence between intermediate representations of an input to the
  labels.
\item
  Parameter attribution: An interpretable model should allow
  correspondence between parameters of the model and the input-label
  decision making process. That is, it should be evident that
  \emph{which} parameters act on \emph{which} parts of inputs to produce
  \emph{which} labels. Again, the number of such parameters and
  inputs-parts, that would be attributed could be reduced for
  simplification.
\end{enumerate}

For any model, interpretation should result in some attribution from the
parameters of the model towards the predictions made by it. This is
separate from \emph{distilling} a network into a more explainable model
like a decision tree \cite{frosst2017distilling}, which aims more
towards converting the model to a simpler one, while allowing for
significant accuracy loss. Model interpretation (and consequent
simplification) however, does not attempt to significantly alter the
model and largely attempts to maintain the model accuracy.

In the context of CNNs, we can look to interpret them on the basis of
parts of images and the filter banks acting on them; as deep CNNs
contain banks of convolutional filters stacked one after the other.
These filters which provide a locally linear weighted average of the
image signal can be considered for interpretability as instead of
individual weights they act in tandem on the input and intermediate
representations.

\hypertarget{cnn-interpretability}{%
\subsection{CNN Interpretability}\label{cnn-interpretability}}

There is sufficient prior work on inspecting and transferring the
filters in a Deep CNN. A detailed survey is given in
\cite{zhang2020survey_a}. Earlier works like
\cite{zeiler2014visualizing}, \cite{szegedy2014intriguing},
\cite{simonyan2014deep} analyzed the filters in intermediate layers and
found that the features from pretrained neural networks can be used for
other tasks.

Later work has moved more towards class attribution and aligning
features to semantic concepts. Gonazelz-Garcia et al.
\cite{gonzalezgarcia2017semantic} and Bau et al. \cite{bau2017network}
analyze the activation maps of CNNs and discover that the final layers
of the networks align closer to semantic concepts than the early layers
which were more attuned to detecting texture and color.

Network Dissection \cite{bau2017network} attempts to associate the
activation maps of CNNs with semantic concepts. Bach et. al.
\cite{bach2015pixelwise} introduce the concept of \emph{Relevance
Propagation} and visualize the contributions of individual pixels in an
image. Hendricks et. al. \cite{hendricks2016generating} use a
Reinforcement Learning based loss function to provide natural language
explanations of the model's predictions. Guillame and Bengio
\cite{alain2017understanding} integrate separate linear classifiers at
each layer to understand the model's decision making process. Zhang et.
al. \cite{zhang2018interpretable} build templates specific to parts and
incorporate that during the training of the CNN to generate
interpretable feature maps. Liang et. al. \cite{liang2020training} use
sparsity to find out the important filters in a CNN, though instead of
going towards a parameter level sparsity like dropout or quantization
\cite{yang2019quantization}, they look at one Convolutional Filter as a
single parameter. Some other approaches like
\cite{zhang2018interpreting} and \cite{zhang2019interpreting} attempt
to estimate the contributions of the filters via modeling them as a
graphical model. The concept of \emph{Class Activation Maps} is
described in \cite{zhou2016learning}, \cite{selvaraju2017gradcam}.
which try to identify discriminative regions of CNNs, while
\cite{wickramanayake2019flex} and
\cite{wickramanayake2021comprehensible} aim for linguistic descriptions
of model explanations.

A parallel line of investigation has studied feature importance in CNNs
and sparse CNNs. Liu et. al. \cite{liu2015sparse} use sparse low rank
decompositions in pretrained CNNs. Li et. al. \cite{li2017pruning_a}
use \(\ell_1\) norm of filters to rank the convolutional filters as a
whole and prune the less important ones. Kumar et al.
\cite{kumar2021pruning} also use \(\ell_1\) norm with a \emph{capped}
\(\ell_1\) norm to formulate classification with CNNs as a \emph{lasso}
like problem. Lin et al. \cite{lin2020toward} introduce a
\emph{structured sparsity regularization}. Li et al.
\cite{li2019exploiting} introduce a \emph{kernel sparsity and entropy}
(KSE) measure which quantifies both sparsity and diversity of the
convolution kernels.

The notion of \emph{class specificity} is explored in Wang et al.
\cite{wang2018learning}. They try to increase accuracy with filters
that capture \emph{class-specific} discriminative patches. These
\emph{class-specific} filters are closely related to earlier work in
Jiang et al. \cite{jiang2017learning}, who introduce the concept of
\emph{Label Consistent Neural Networks} to learn features which they
claim alleviate gradient vanishing and leads to faster convergence.
Liang et al. \cite{liang2020training} try to use sparsity to find out
the important filters which they claim are also \emph{class specific}.

Our work takes inspiration from \cite{li2017pruning_a},
\cite{wang2018learning} and \cite{liang2020training}. We combine the
ideas of \emph{class specificity} \cite{wang2018learning},
\cite{liang2020training} and \(\ell_1\) norm based filter importance
\cite{li2017pruning_a} to arrive at a method for identifying the
\(k\) most influential features based on an \(\ell_1\) norm importance
criterion in the final and penultimate layers of a CNN and demonstrate
its efficacy with experiments on various CNNs.

\hypertarget{our-contributions}{%
\subsection{Our Contributions}\label{our-contributions}}

Our contributions are following:

\begin{enumerate}
\def\labelenumi{\arabic{enumi}.}
\tightlist
\item
  We show that only a few filters \emph{per class} are needed to make a
  decision for a deep CNN.
\item
  We provide an algorithm to obtain those filters from \emph{any
  pre-trained} network with a single fully connected layer.
\item
  We demonstrate the relation between depth and filter disentanglement
  in CNNs and show that deeper networks lead to lower dimensional
  representations in the final layer.
\item
  We show that these filters have a greater correspondence to objects
  within the image and are critical to classification for that class.
\end{enumerate}

The rest of the paper is organized as follows: We discuss CNNs and the
concept \emph{class specific features} in Section \ref{background}. We
provide an overview of analysis of techniques on final layer and
\emph{class specific features} therein, in Section
\ref{influential-features}. We describe the experimental details and
results in Section \ref{experiments-and-results}. We discuss
implications and future scope in Section
\ref{discussion-and-future-work}.

\hypertarget{background}{%
\section{Background}\label{background}}

We discuss CNNs, filters and layers in the next sections and establish
notation which will aid us in our describing our methods.

\hypertarget{notation}{%
\subsection{Notation}\label{notation}}

Let \(\{(\mathcal{X,Y})\}\) be the set of data tuples. An instance of
the data
\((X,y) \in \mathcal{\{(X,Y)\}}\) is a tuple of (image, label) with \(X \in \mathbbm{R}^{c \times h \times w}\)
where \(c,h,w\) are the channels, height and width of the image
respectively.
\(y\) is an integer response variable representing one of the \(n\)
classes, \(y \in \mathbbm{Z}^{+}, 0 \leq y < n\). \(y\) can also be
represented as an index vector indexing the
\(i^{th}\) class, \(y \in \{0,1\}^n: y = i \equiv \{0,0,...,1,...,0\}\).

Formally a Deep Neural Network is a set of weights which act as a
sequence of operators on input \(X\), such that,
\(y = \mathcal{A} W^d (... \mathcal{A} W^1 (\mathcal{A} W^0 (X)))\)
where \(W^d\) is a weight tensor at depth \(d\) and \(\mathcal{A}\) is
an element-wise function also known as an \emph{activation function}.
While the activation function
\(\mathcal{A}\) need not be same for each weight \(W^d\), for CNNs that
we consider only RELU \(\max(0, x)\) is used. An exception is the
weights at the end of the network where a softmax
\(\sigma(x) = \exp(x_j) / \sum \exp(x_j)\) is used.

A \emph{layer} of a network is a single weight+activation operation
\(O^d = \mathcal{A}(W^d(I^d))\) where \(I^d,W^d,O^d\) are input, weights and output at depth \(d\).
We will denote the layer at depth \(d\) as \(L^d\).

\hypertarget{cnns}{%
\subsubsection{CNNs}\label{cnns}}

A Convolutional Network consists of filter banks of \emph{convolution}
(or cross correlation) filters which are square matrices of odd rank
acting on the input with an element-wise activation function on the
output. Such an output \(O^d\) at layer \(d\) is called a \emph{feature
map} at \(d^{th}\) layer. Filters are the fundamental unit for a CNN and
it is convenient to represent and analyze a CNN as operations due to the
filters. For our purposes, we will denote the
\(j^{th}\) filter for layer \(i\) as \(L^i_j\).

Modern Deep CNNs rely heavily on RELU and Batch Normalization
\cite{ioffe2015batch} in the intermediate layers. Batch Normalization
and RELU are performed after the Convolution operator and the entire
unit can be considered as a single operator. A convolution operation of
a filter \(w\) of size \(k \times k\) on an input image matrix
\(I\) of size \(h \times w\) is defined as: \(\operatorname{Conv}_{w_{k \times k}}: I \rightarrow O\)
where
\(O_{i,j} = {\displaystyle\sum_{l=-k^{'}}^{k^{'}}} I_{i-k,j-k} w_{k-l,k-l}\) and \(k^{'} = \lfloor k/2 \rfloor\)

\begin{center}
  \begin{figure}
    \centering
    \includegraphics[height=2in]{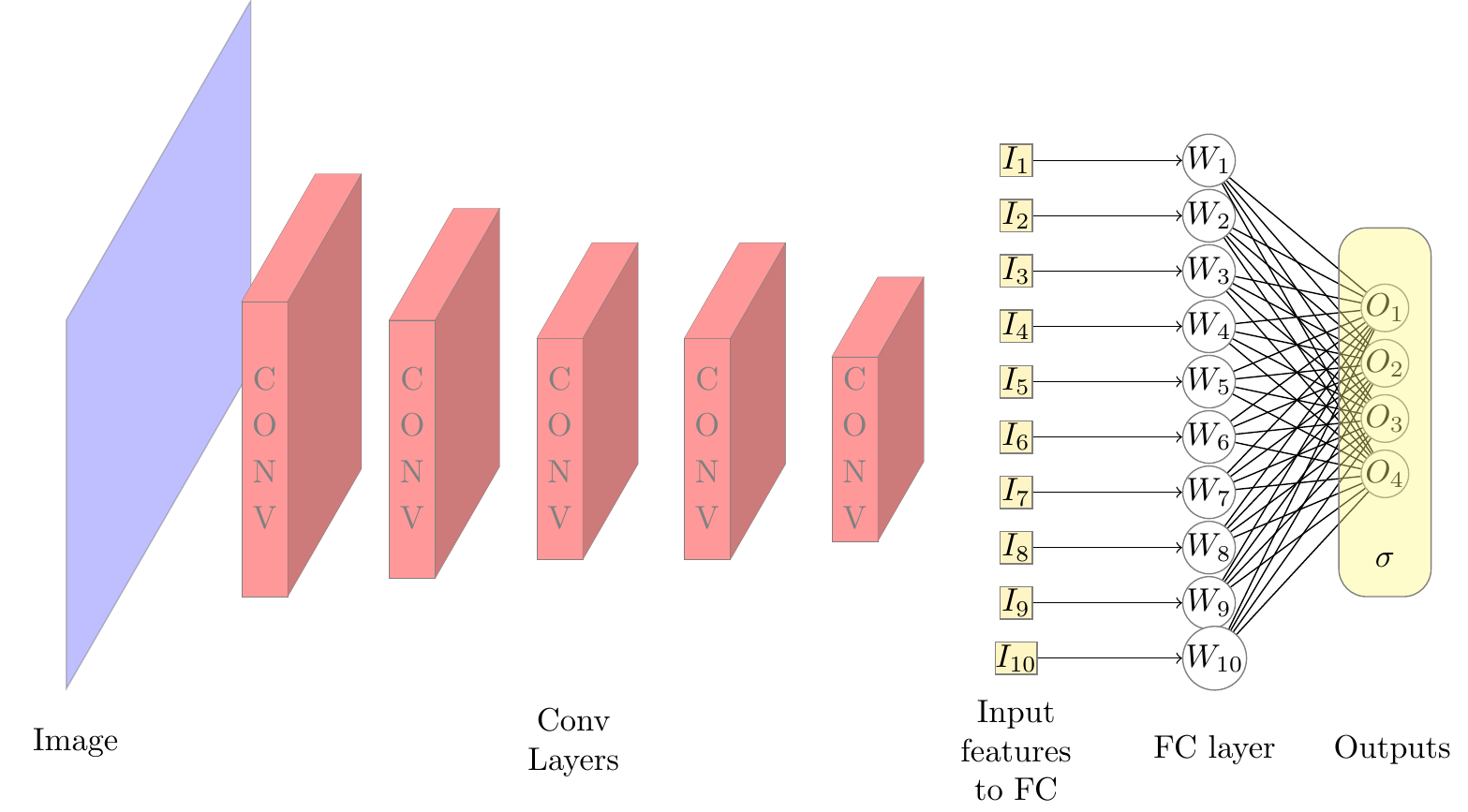}
    \caption{CNN with Fully Connected Final Layer. An image is input at the
      \emph{head} of the network where it travels sequential through the
      convolutional layers (denoted by CONV). At the final layer (FC layer) the
      input coming from the final convolutional layer is flatted to a
      vector. The predictions are done by applying softmax $\sigma$ elementwise
      to the outputs.}
    \label{fig:CNN}
  \end{figure}
\end{center}

Let \(I^d\) be the input at the final layer of the CNN with \emph{total
depth}
\(d\). The output at the final layer is then \(O^d = \sigma(WI^{d})\),
where
\(\sigma\) is the \(\operatorname{softmax}\) operator. \(\sum_i O^d_i = 1\)
because of softmax and thus it can be interpreted as a probability
distribution over \(y\). The probability of each class
\(y_i\) is given by \(O^d_i\) and the most probable label
\(\hat{y}\) is given by \(\hat{y} = \operatorname{argmax}(O^d)\). See
Fig \ref{fig:CNN} for an illustration.

\vspace{-2em}

For further details, We refer the reader to
\cite{krizhevsky2012imagenet}, \cite{szegedy2015going} and
\cite{he2016deep} for internals of CNNs.

\hypertarget{filter-disentanglement-and-label-consistency}{%
\subsection{Filter Disentanglement and Label
Consistency}\label{filter-disentanglement-and-label-consistency}}

As mentioned earlier, the sheer number of features and the layers in
CNNs makes them very hard to interpret. One approach towards their
interpretation is \emph{Filter Disentanglement}. Filter Disentanglement
refers to the fact that the filters in a CNN should represent separate
concepts. As concepts in a network are hard to identify, we can instead
use the attribution of a filter towards the prediction of a label as a
surrogate measure of disentanglement. \emph{Ideally} every filter should
be responsible for detection of a particular pattern in the image, but
that is difficult in practice. In particular at the bottom layers the
filters are highly entangled and learn very generic features
\cite{bau2017network}. As we move up the network, at the top layers
they tend to be less entangled but not entirely so.

As in Section \ref{background}, let the set of filters for a CNN at
depth \(d\) constitute the layer
\(L^d\). Now consider a single label \(y \in \mathcal{Y}\). Let
\(\exists \text{ } L^d_{J_y} \subset L^d\) where \(J_y\) is an index set such that, \(L^d_{J_y}\)
\emph{alone} is responsible for the prediction of label \(y\).

For such a set
\(J_y\), we define the \emph{influence} relation \(\prec\) for the probability \(P(y^d)\)
of a label \(y\) at depth \(d\) of a CNN as: \(P(y^d) \prec L^d_{J_y}\).
We say that the features
\(L^d_{J_y}\) are \emph{influential} at depth \(d\) for the prediction
of label \(y\).

We will focus only on the final layer so we can remove the superscript
\(d\) without ambiguity. To enforce disentanglement and considering only
the final layer we should ideally have,

\begin{enumerate}
\def\labelenumi{\arabic{enumi}.}
\tightlist
\item
  \(|L_{J_y}| \ll |L|, \forall y\) and,
\item
  \(L_{j_{y_1}} \neq L_{j_{y_2}}, \forall y_1,y_2 \in \mathcal{Y}\)
\end{enumerate}

That is, the filters at depth \(d\) for each class \(y\) should be
disjoint. However, as we mentioned above, that is not practically
feasible. Instead we can hope to find:

\begin{enumerate}
\def\labelenumi{\arabic{enumi}.}
\tightlist
\item
  \(|L_{J_y}| \ll |L|, \forall y\)
\item
  \(\underset{J_{y_1},J_{y_2}}{\operatorname{argmin}}(L_{J_{y_1}} \cap L_{J_{y_2}}), \forall y_1,y_2 \in \mathcal{Y}\)
\end{enumerate}

Or we should seek index sets \(J_{y_i}\) such that there is minimal
overlap between two classes. These \emph{class-specific} filters are
closely related to \cite{jiang2017learning}, except that
\cite{jiang2017learning} associate a \emph{neuron} (or weight) with a
label while we associate filters. The \emph{class-specific} filters
discussed in Liang et. al. \cite{liang2020training} are more similar to
these. However, unlike \cite{liang2020training} and
\cite{wang2018learning} our method does not require supervision and can
be used for any pretrained network. And, although we use \(\ell_1\) norm
to identify filters, our method identifies \emph{class specific} filters
and decomposes the final layer while the method in
\cite{li2017pruning_a} does not.

\hypertarget{influential-features}{%
\section{Influential Features}\label{influential-features}}

The penultimate and final layers are of particular interest in feature
attribution in CNNs as they contain the final representation of the
image and directly lead to the prediction of the label, while the lower
layers learn the filters which respond to textures and lower level
features \cite{bau2017network}. Here we describe the various factors in
determining the \emph{influential features} in the final layer.

\hypertarget{ell_1-norm-for-feature-importance}{%
\subsection{\texorpdfstring{\(\ell_1\) norm for Feature
Importance}{\textbackslash ell\_1 norm for Feature Importance}}\label{ell_1-norm-for-feature-importance}}

Previously \cite{li2017pruning_a} and \cite{liang2020training} have
both used \(\ell_1\) norm for estimating filter importance. However,
earlier works such as the two above, do not explicitly discover a lower
dimensional decision surface. Following \cite{li2017pruning_a} and
\cite{liang2020training} we look at the \(\ell_1\) norm of the
resulting \emph{features per class} as the primary differentiating
factor.

One approach would be to check the \(\ell_1\) norm of \emph{weights} in
the final and pre final layers which is similar to
\cite{liang2020training}. However we can note that the \(\ell_1\) norm
of the weights will not vary for each class so it is much harder to
identify \emph{class specific} features with that. In our initial
experiments also, this hypothesis was verified.

Another key observation we make is that all the features in the entire
CNN are \(\geq 0\) because of RELU. In effect the feature
vectors/tensors are positive semidefinite, which means that the
\(\ell_1\) norm of each feature directly contributes to the
classification output of the final layer. The weights on the other hand
are roughly \(50\% \geq 0\) and \(50\% \leq 0\) and therefore it's
easier to quantify feature importance than weight importance with
\(\ell_1\) norm.

\hypertarget{separating-outputs-for-each-class}{%
\subsection{Separating Outputs for each
Class}\label{separating-outputs-for-each-class}}

A second insight we had is that the final layer output \emph{for each
class} is the result of a dot product between features for that class
and weights for that class. Therefore the final layer can be thought of
as \(c\) separate dot products, where \(c = |\mathcal{Y}|\) as earlier
is the number of classes. Hence, selecting certain features \emph{for
each class} will not affect the output for the other.

Formally, \(P(y|X) = O^d = \sigma(WI^{d})\) as in Section
\ref{background}, where
\(O^d \in \mathbbm{R}^n\) and \(\sum_i O^d_i = 1\). If we omit
\(d\) and \(X\) for simplicity, then probability for \(i^{th}\) class,
\(P(y=i)\) is the value of \(i^{th}\) component of the output from the
final layer, which is \(O_i\).

Now the final layer is a single matrix, so \(O_i\) is simply the dot
product of \(i^{th}\) row of the weight matrix \(W\) with the input
feature vector \(I\). If \(m\) is the \emph{width} of the final layer
then,
\(O_i = W_i \cdot I\), where \(W \in \mathbbm{R}^{m \times n}\) and \(W_i, I \in \mathbbm{R}^{m}\).

Now, let
\(\mathbf{w}_{k,i}\) be the \(k\) dimensional subspace of \(W_i\) for
the index \(i\), that is, \(\mathbf{w}_{ki} \subset W_i\),
\(\mathbf{w}_{k,i} \in \mathbbm{R}^k, k \ll m\). Then we define the
probability for label \(i\) at final layer with reduced dimension \(k\)
as \(P \left( y^k=i \right) = \sum \mathbf{w}_{k,i} I_k\).

Recall that the predicted label with the full \emph{width} is
\(\hat{y} = \operatorname{argmax}(y)\). We define a prediction with
reduced dimension \(k\) as \(\hat{y}_k = \operatorname{argmax}(y_k)\).
The goal then is to find such \(\mathbf{w}_{k,i}\) for each class \(i\),
such that the difference in the predictions
\(d(\hat{y}_k, \hat{y})\) is minimized, where \(d\) is some metric.

That is, \(\mathbf{w}_k\) are the truncated weights which minimize the
difference between the predicted labels at \emph{width} \(k\) and
predicted labels at full \emph{width} \(n\) for each class.

\hypertarget{finding-the-influential-features}{%
\subsection{Finding the Influential
Features}\label{finding-the-influential-features}}

We note that finding the \emph{class specific influential features}
\(\mathbf{w}_{k,i}\) is non-trivial as the exact subset of the weights
cannot be known easily and an exhaustive search is of exponential
complexity. However, as we mention in Section
\ref{ell_1-norm-for-feature-importance} \(\ell_1\) norm of the features
can help us in guiding towards the correct set of filters.

In our experiments we found that although selecting weights by top
\(\ell_1\) norm would result in \emph{weights attribution}, it would not
result in \emph{class specific features} attribution. Instead searching
for \emph{topk} \emph{features per class} with \(\ell_1\) norm gave us
better results. Even combining \emph{topk} \emph{filters per class} with
\emph{topk} weights led to poorer results than with \emph{topk} features
per class.

A problem though, is that the features per instance vary across a single
class, and the label of the instance cannot be known in advance. However
a \emph{topk} selection from \emph{histogram of topk features for all
instances} in a given class had mass concentrated around a few points,
which corresponded to such \emph{influential features}. We give the
details of the algorithm in the next section.

\hypertarget{algorithm}{%
\subsection{Algorithm}\label{algorithm}}

Here we discuss the algorithm to obtain the \(k\) most \emph{influential
features} for each class from a pretrained CNN. For the algorithm
\ref{algo:influential} below, \({\mathcal{I}}\) are the set of features
at final layer for all classes, where we omit the layer superscript used
earlier for simplicity. Denote the set of features for label
\(y\) by \(I_y\). We want to obtain the mapping
\(\mathbb{I}: y \rightarrow I_y\), such that, \(\mathbb{I}(y)\) gives
the most influential features for class \(y\). The parameters for the
algorithm are \(k_1, k_2 \in \mathbbm{Z}^{+}\) which define the initial
and subsequent feature selection, which we describe below.

We proceed by noting top \(k_1\) features by \(\ell_1\) norm of each
data instance for \emph{each class} at the pre-final layer and select
the indices in a set
\(I^{topk_1}_y\) for class \(y\). We set \(k_1 \ll m\) where \(m\) is
the dimension of the features at the pre-final layer (and hence, also
the dimension of the final layer). E.g., if for a 64 dimensional feature
vector, let the top 5 components sorted by
\(\ell_1\) norm occur at indices \(<3, 5, 14, 18, 28>\). Then the index
set for \(k_1 = 5\) for the entire class is aggregation
\(\biguplus I^{top5}_y\) of all such sets, where \(\biguplus\) is an
aggregation operator, e.g.~\(\biguplus <3, 5> <3, 14> = <3, 3, 5, 14>\)

\begin{center}
  \begin{figure*}
    \includegraphics[height=3in]{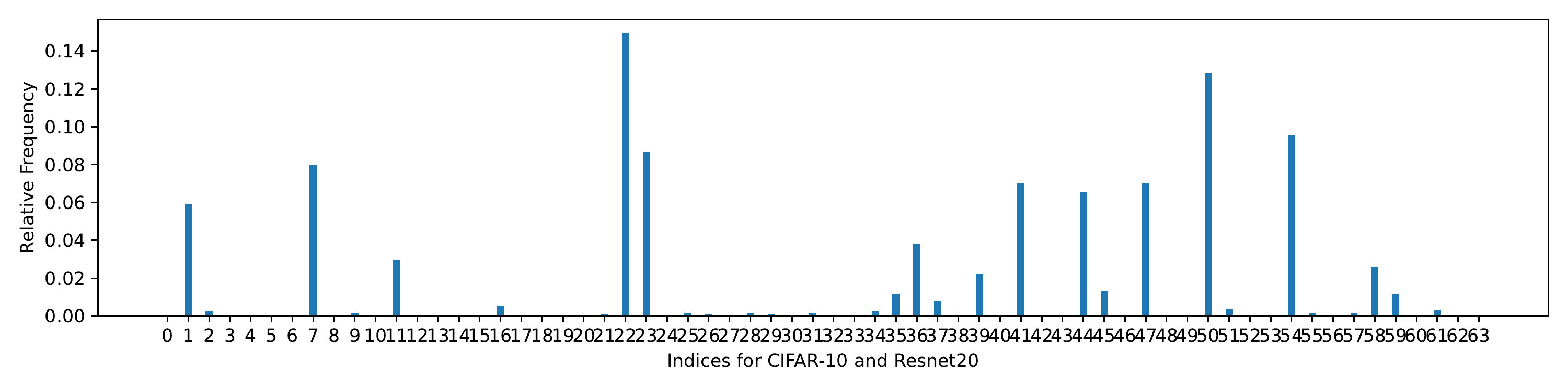}
    \caption{Histogram of relative frequencies indices for CIFAR-10 and Resnet20. The mass concentrates around a few indices.}
    \label{fig:hist_cifar}
  \end{figure*}
\end{center}

The set \(\biguplus I^{topk}_y\) then denotes all the occurrences of a
particular dimension (or index) of a feature in \(topk\) \(\ell_1\) norm
set, for a class in pre-final layer. The frequency distribution for one
such class for Resnet20 is given in Fig. \ref{fig:hist_cifar}. From that
histogram we then select the top \(k_2\) most frequent indices.

We also experimented with
\(k_1\) which covers a certain percentage (say \(90%
\)) of contribution of the filters. However, we found that a value of 5
for CIFAR-10 \cite{krizhevsky2009learning} gave us \(90%
\) coverage, which we then chose to set for all our experiments as that
is faster to implement.

For Imagenet \cite{deng2009imagenet} the number of filters which
contributed over \(90%\) by \(\ell_1
\) norm was very large, and we restricted ourselves to top 50. Even
though the intial accuracy after decomposition is lower, we are able to
retrain the network back close within \(1%
\) of original accuracy.

\begin{algorithm}
  \caption{Extract Influential Features}\label{algo:influential}
  \begin{algorithmic}[1]
    \State \textbf{Input:}\hspace{2pt} $\mathcal{I}$, $\mathbb{I}$, $k_1$, $k_2$
    \State \textbf{Output:}\hspace{2pt} Mapping $\mathbb{I}$ of Influential Features
    % \Procedure {HIST}{$I$}
    % \State
    % \EndProcedure
    \Procedure {get\_indices}{$\mathcal{I},y$}
    \For {$I_y \leftarrow \mathcal{I}$}
    \For {$i_y \leftarrow {I_y}$}
    \State $I^{topk_1}_y = I^{topk_1}_y \biguplus \left (\operatorname{sorted}_k(i_y) \right )$ w.r.t. $\ell_1$ norm
    \EndFor
    \State $\mathbb{I}_y \leftarrow topk_2 \left (\operatorname{HIST}(I^{topk_1}_y) \right )$
    \State $\mathbb{I} = \mathbb{I} \biguplus \left (y,\mathbb{I}_y \right )$ \Comment{$\left (y,\mathbb{I}_y \right )$ \text{ is a tuple of } $y$ \text{ and } $\mathbb{I}_y$}
    \EndFor
    \EndProcedure
  \end{algorithmic}
\end{algorithm}

HIST in algorithm \ref{algo:influential} refers to the histogram of
frequencies of each index.

\vspace{-4em}

\hypertarget{experiments-and-results}{%
\section{Experiments and Results}\label{experiments-and-results}}

We conducted some initial experiments on Class Specific Gates (CSG)
\cite{liang2020training} and we discuss them in
\ref{class-specific-filters} as to illustrate our argument. We then
analyze and discuss the efficacy of our proposed \emph{influential
features} in \ref{influential-features-1}. As mentioned earlier, we take
inspiration from \cite{li2017pruning_a} and \cite{liang2020training}
and combine class specificity with \(\ell_1\) norm importance. We note
that while \cite{liang2020training} is a supervised method, their
approach is closest to ours in principle. \cite{li2017pruning_a} is
unsupervised but their aim is to induce sparsity and not
interpretability. Also their approach doesn't lead to explicit class
specific disentanglement. We demonstrate that \textbf{our unsupervised
approach} gives as a) good if not better results than
\cite{liang2020training} b) leads to explicit filter disentanglement c)
reduces computational cost in the final layer.

\hypertarget{class-specific-filters}{%
\subsection{Class Specific Filters}\label{class-specific-filters}}

We first discuss experiments with \emph{class specific gates} (CSG).
Liang et al. \cite{liang2020training} had proposed to learn a matrix
\(\operatorname{CSG} \in \mathbbm{R}^{m \times n}\) where has
\(m\) is the dimension of final layer and \(c = |\mathcal{Y}|\) is the
number of classes. See \cite{liang2020training} for details of the
training procedure. One important issue to consider with
\cite{liang2020training} is that the label CSG matrix would not be
available when doing inference as the labels for new data is not known.
Hence it can only be used to interpret a CNN on a given dataset and
cannot be used for any new data.

While \cite{liang2020training} have released part of the code for their
experiments, they did not release the CSG learning code, especially CSG
for CIFAR-10 and Imagenet\footnote{See
  \url{https://github.com/hyliang96/CSGCNN}}. We implement CSG and
summarize the results in Table \ref{tab:csg_ours}.

STD output in the Table \ref{tab:csg_ours} is
\(O^{STD}_i = O^d_i = \sigma(WI^{d})_i\), which is the same as CNN
output without a CSG matrix. Output using CSG matrix is
\(O^{CSG}_i = \sigma(W (I^{d} \odot \operatorname{CSG}_{i}))\), where \(\odot\)
denotes the Hadamard or element-wise product. We note that the CSG
accuracy is similar to the STD accuracy. CSG and STD are CSG and STD
output paths respectively.

\begin{table}
  \centering
  \caption{CSG Experiments Conducted by us}
  \begin{tabular}{|l|l|l|l|}
    \hline
    Dataset & Model & Training & Accuracy \\
    \hline
            & Resnet20 & CSG & 0.8809 \\
            & & STD & 0.8809 \\
    CIFAR-10 & Resnet34 & CSG & 0.8407 \\
            & & STD & 0.8407 \\
    \hline
    Tiny & Resnet18 & CSG & 0.3633 \\
    Imagenet & & STD & 0.3641 \\
    200 & Resnet34 & CSG & 0.3794 \\
            & & STD & 0.3852 \\
    \hline
  \end{tabular}
  \label{tab:csg_ours}
\end{table}

\hypertarget{influential-features-1}{%
\subsection{Influential Features}\label{influential-features-1}}

We have performed experiments to obtain \emph{influential filters}
according to Algorithm \ref{algo:influential} on families of models a)
Resnet b) Densenet c) Efficientnet to demonstrate that our method can
work for any CNN. We've used standard datasets CIFAR-10
\cite{krizhevsky2009learning} and Imagenet \cite{deng2009imagenet} in
our experiments.

Resnet \cite{he2016deep} are the most popular variants of CNNs because
of their lower computational cost and good generalization in spite of
it. Densenets \cite{huang2017densely} are another popular architecture
which connect each layer to every other layer. Efficientnets
\cite{tan2019efficientnet} are recent models which are developed with
Neural Architecture Search \cite{zoph2017neural} and aim to reduce the
computational cost while preserving accuracy. The reader is referred to
the respective papers for the details of the models.

Resnet family of models is our primary focus for retraining,
visualization and criticality experiments. We calculate and compare
\emph{influential features} for all the other models.

CIFAR-10 consists of 50,000 training images for 10 classes, i.e.,
\(c = |\mathcal{Y}| = 10\), of size \(32 \times 32\) and 10,000
validation images. Imagenet has 1,000,000 images of varying sizes with
\(c = 1000\) but while training they are resized to \(224 \times 224\).
The validation split has 50,000 images for Imagenet. For both the
datasets and all the models we determine
\(k_1, k_2\) and \(\biguplus I^{topk}\) only from the training set. The
results are then calculated on the validation set.

\hypertarget{effect-of-k_1-and-k_2}{%
\subsubsection{\texorpdfstring{Effect of
\(k_1\) and \(k_2\)}{Effect of k\_1\textbackslash) and \textbackslash(k\_2}}\label{effect-of-k_1-and-k_2}}

\begin{center}
  \begin{figure}
    \centering
    \includegraphics[height=2.2in]{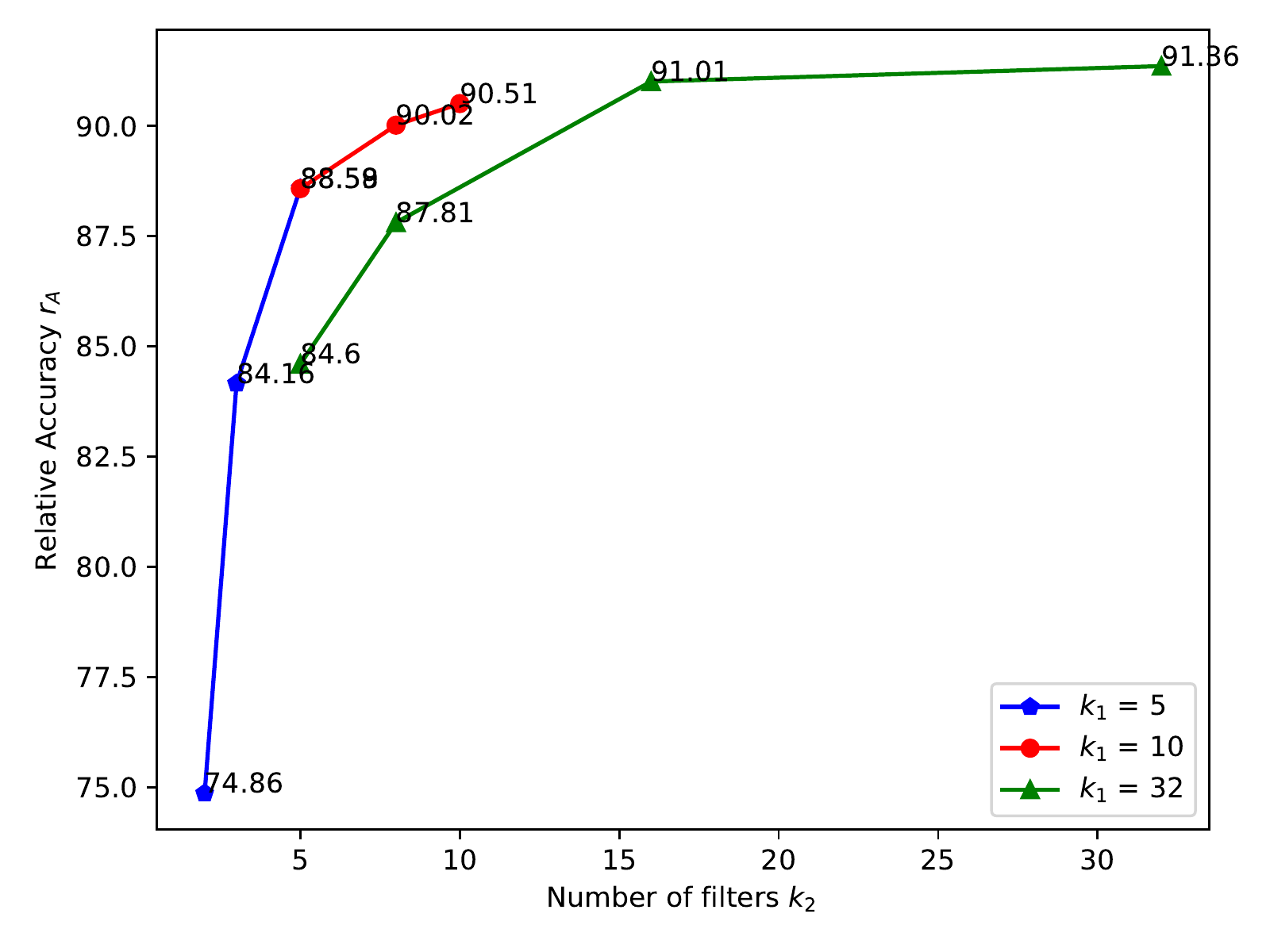}
    \caption{Effect of $k_1$ and $k_2$ on Accuracies for Resnet20 and CIFAR-10}
    \label{fig:k1k2}
  \end{figure}
\end{center}

We conduct a detailed study on Resnet20 model and CIFAR-10 dataset to
analyze the effect of values of \(k_1\) and \(k_2\) on the resulting
model. For the Resnet20 model, our pretrained model had accuracy
\(91.17%
\). We measure the efficacy of the resulting decomposed final layer as
the ratio \(r_A = \frac{A_d}{A_f}\), where \(A_d\), \(A_f\) are the
decomposed and full width accuracies respectively. See Fig
\ref{fig:k1k2} for the results.

\vspace{-3em}

We observe overall that reducing the dimensionality by lowering the
number of input features to the final layer also results in a reduction
of accuracy. This is to be expected as we are losing information at the
last layer. We note that even for \(k_2 = 3\), we get \(84.16%
\) accuracy which is pretty good for only 3 filters per class. This
confirms that the decision surface lies on a much lower dimensional
subspace than the dimension of the final layer.

Another thing we can note is that, for a given value of \(k_1\) the best
results are obtained by setting \(k_2 = k_1\). Also, if initial filter
selection \(k_1\) is very large, then equivalent \(k_2\) setting leads
to lower performance as compared to a lower \(k_1\).

\hypertarget{effect-of-depth}{%
\subsubsection{Effect of Depth}\label{effect-of-depth}}

Here we discuss the results of experiments on various other models
including variants of Resnet. For all the following experiments we use
Imagenet Dataset and set \(k_1 = k_2 = 50\). Two important metrics in
experiments with depth are
\(k_2/n\) and number of filters per class \(k_2/c\). For all variants of
Resnet \(k_2/n, k_2/c\) and dimension of final layer \(n\) remain the
same. Table \ref{tab:resnet_depth} summarizes the results for Resnet
models.

\begin{table}
  \centering
  \caption{Effect of depth on Resnet variants}
  \begin{tabular}{|c|c|c|c|}
    \hline
    & Resnet50 & Resnet101 & Resnet152 \\
    \hline
    Final Dim $n$ & \multicolumn{3}{|c|}{2048} \\
    \hline
    $k_2/c$ & \multicolumn{3}{|c|}{0.05} \\
    \hline
    $k_2/n$ & \multicolumn{3}{|c|} {0.0244141} \\
    \hline
    $A_d$ & 0.64792 & 0.68164 & 0.69346 \\
    \hline
    $A_f$ & 0.74548 & 0.75986 & 0.77014 \\
    \hline
    $r_A$ & 0.869131 & 0.89706 & 0.900434 \\
    \hline
  \end{tabular}
  \label{tab:resnet_depth}
\end{table}

We can see that the relative accuracy \(r_A\) drops for Imagenet in
these variants compared to Resnet20, but the number of filters per class
\(k_2/c\) is also lower at 0.05 as compared to 0.5 for Resnet20 which is
an order of magnitude. \(k_2/n\) is also lower at 0.024 as compared to
0.078. This is due to the much higher number of classes in the Imagenet
dataset. Apart from that the effect of \(r_A\) on depth is clear as it
increases monotonically with increasing depth.

We also evaluate on Wide Resnets \cite{zagoruyko2016wide} which contain
a greater number of convolution filters per layer as compared to
standard Resnet models. We get a much higher relative accuracy \(r_A\)
on these as compared to standard Resnets for the same number of layers
(see Table \ref{tab:wide_resnet}). We suspect that the greater number of
convolutional filters help in disentanglement as there are more filters
per class in the previous layers.

\begin{table}
  \centering
  \caption{Effect of depth on Wide Resnet}
  \begin{tabular}{|l|l|l|}
    \hline
    & Wide Resnet 50 & Wide Resnet 101 \\
    \hline
    Final Dim $n$ & \multicolumn{2}{|c|}{ 2048} \\
    \hline
    $k_2/c$ & \multicolumn{2}{c|}{0.05} \\
    \hline
    $k_2/n$ & \multicolumn{2}{c|}{0.0244141} \\
    \hline
    $A_d$ & 0.75008 & 0.75988 \\
    \hline
    $A_f$ & 0.77256 & 0.77908 \\
    \hline
    $r_A$ & 0.970902 & 0.975356 \\
    \hline
  \end{tabular}
  \label{tab:wide_resnet}
\end{table}

For all the models we can see that across a class of models \(r_A\)
increases with depth. Only EfficientNet deviates from this behaviour
which we discuss later. Densenets \cite{huang2017densely} have
different \(n\) and hence \(k_2/n\) differs for them. For Densenets we
compare Densenet121 with Densenet169 and Densenet161 with Densenet201,
as the two pairs have closer \(n\) and hence \(k_2/n\).

\begin{table}
  \centering
  \caption{Effect of depth on Densenet 121 and 169}
  \begin{tabular}{|l|l|l|}
    \hline
    & Densenet121 & Densenet169 \\
    \hline
    Final Dim $n$ & \multicolumn{2}{|c|}{ 2048} \\
    \hline
    $k_2/c$ & \multicolumn{2}{c|}{0.05} \\
    \hline
    $k_2/n$ & 0.0488281 & 0.0300481 \\
    \hline
    $A_d$ & 0.6372 & 0.66328 \\
    \hline
    $A_f$ & 0.71956 & 0.73754 \\
    \hline
    $r_A$ & 0.885541 & 0.899314 \\
    \hline
  \end{tabular}
  \label{tab:densenet_depth_1}
\end{table}

\begin{table}
  \centering
  \caption{Effect of depth on Densenet 121 and 169}
  \begin{tabular}{|l|l|l|}
    \hline
    & Densenet161 & Densenet201 \\
    \hline
    Final Dim $n$ & \multicolumn{2}{|c|}{ 2048} \\
    \hline
    $k_2/c$ & \multicolumn{2}{c|}{0.05} \\
    \hline
    $k_2/n$ & 0.0226449 & 0.0260417 \\
    \hline
    $A_d$ & 0.6766 & 0.68214 \\
    \hline
    $A_f$ & 0.75268 & 0.7455 \\
    \hline
    $r_A$ & 0.898921 & 0.91501 \\
    \hline
  \end{tabular}
  \label{tab:densenet_depth_2}
\end{table}

\begin{table*}
  \centering
  \caption{Effect of depth on Efficientnet variants}
  \begin{tabular}{|l|l|l|l|l|}
    \hline
    & Efficientnet\_b0 & Efficientnet\_b1 & Efficientnet\_b2 & Efficientnet\_b3 \\
    \hline
    Final Dim $n$ & 1280 & 1280 & 1408 & 1536 \\
    \hline
    Num Layers & 82 & 116 & 116 & 131 \\
    \hline
    $k_2/c$ & 0.05 & 0.05 & 0.05 & 0.05 \\
    \hline
    $k_2/n$ & 0.0390625 & 0.0390625 & 0.0355114 & 0.0325521 \\
    \hline
    $A_d$ & 0.64838 & 0.7046 & 0.6778 & 0.64842 \\
    \hline
    $A_f$ & 0.7609 & 0.76392 & 0.76762 & 0.76928 \\
    \hline
    $r_A$ & 0.852122 & 0.922348 & 0.882989 & 0.842892 \\
    \hline
  \end{tabular}
  \label{tab:efficientnet_depth}
\end{table*}

\begin{center}
  \begin{figure*}
    \centering
    \includegraphics[height=2.2in]{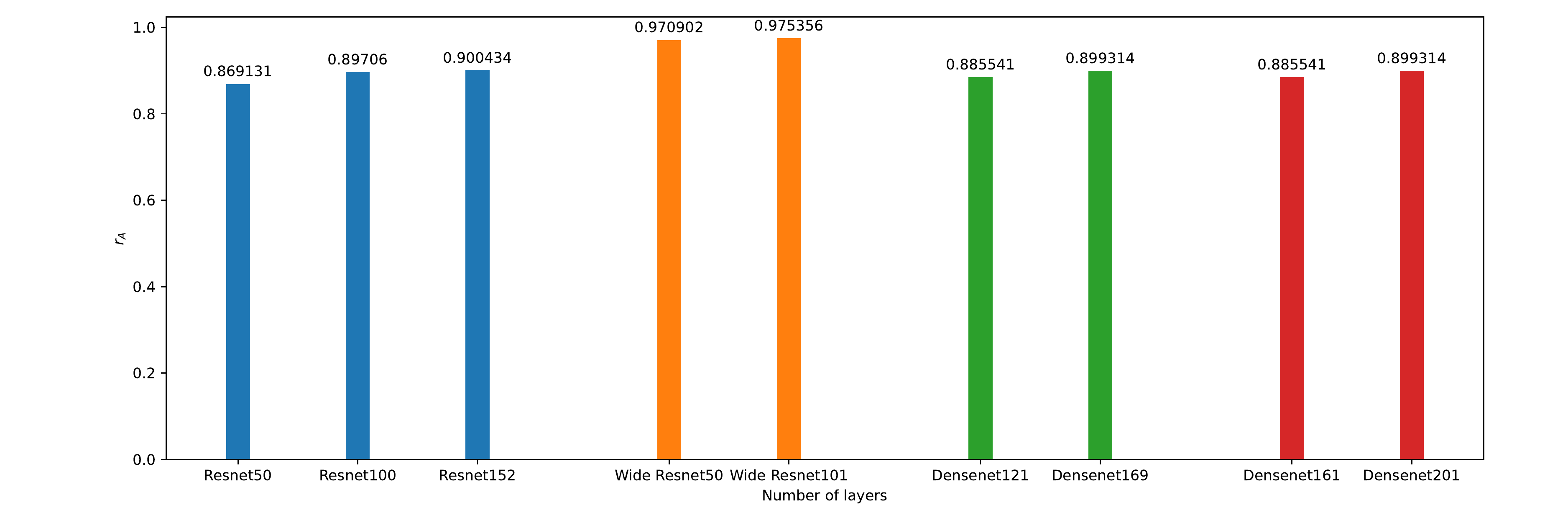}
    \caption{Effect of number of layers (depth) of a network with Relative Accuracy $r_A$.
      We can see that $r_A$ increases with depth for a class of models.}
    \label{fig:depth_comp}
  \end{figure*}
\end{center}

\vspace{-3em}

Coming to Densnets, we can see a similar trend of increasing \(r_A\)
with depth (Tables \ref{tab:densenet_depth_1},
\ref{tab:densenet_depth_2}). The \(r_A\) here is comparable to that of
Resnets but not Wide Resnet variants. The final layer dimension \(n\)
also differs for Densenets so we compare models with a similar
\(n\). We can see that \(r_A\) tends to increase with depth which shows
the effect of depth on filter disentanglement. See Figure
\ref{fig:depth_comp} for a summary of \(r_A\) on all models.

Only with Efficientnets (Table \ref{tab:efficientnet_depth}), which are
NAS based models, do we see a deviation from this pattern as their model
architecture differs from human designed networks. With Efficientnets,
as the number of layers increases from 82 to 116 we see a corresponding
jump in \(r_A\) except Efficientnet\_b2 which has same number of layers
as Efficientnet\_b1 but has a wider final layer, we see a drop in
\(r_A\). Efficientnet\_b3 is a curiosity as that model is both deeper
and wider.

\hypertarget{decomposition-and-prediction-with-influential-features}{%
\subsubsection{Decomposition and Prediction with Influential
Features}\label{decomposition-and-prediction-with-influential-features}}

\begin{center}
  \begin{figure}
    \centering
    \includegraphics[height=2in]{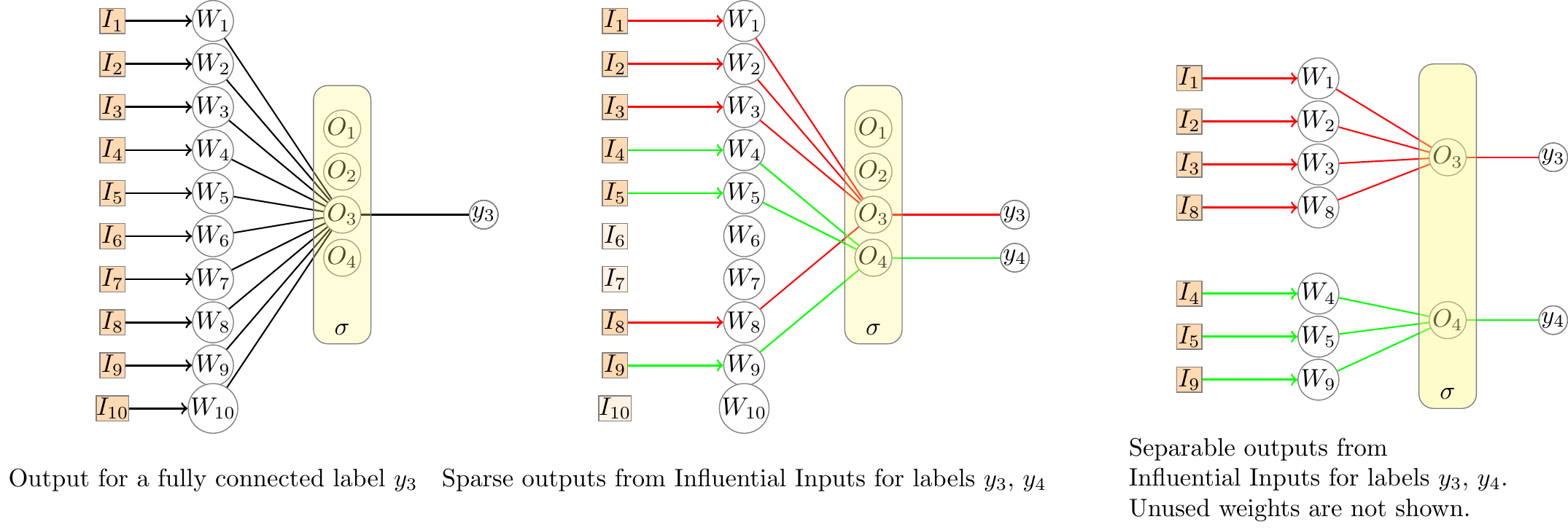}
    \caption{Prediction with influential features for class $y_3$, $y_4$. Only 4
      and 3 input features are required respectively. The final layer can be
      decomposed as shown and the probability for label can be is
      $\sigma(I_{inf}\odot W_{inf})$}
    \label{fig:pred_inf}
  \end{figure}
\end{center}

The computational cost of predicting with \emph{influential features} is
also less as only a few filters are needed for predicting a class.
Because of identification of explicit \emph{class specific filters} we
can decompose the final layer into class specific subspaces. The network
\textbf{can then be retrained to recover the original accuracy}. For
Resnet20 and CIFAR-10 the accuracy drops to 88.59 with \emph{influential
features} with \(k_1 = k_2 = 5\). After that, we can decompose the final
layer so that each class has a separate subspace of features. After
training for one epoch we get accuracy of
\(91.51%\), which is very close to the original accuracy of \(91.71%
\). An illustration of how efficient prediction with \emph{influential
features} works is given in Figure \ref{fig:pred_inf}.

For Resnet50 with Imagenet, after decomposition the accuracy is \(64.7%
\). Unlike with CIFAR-10, it requires around 10 epochs to get close to
the original accuracy of the pretrained Resnet50 and we get \(73.91%
\).

The computational complexity for the final layer is
\(d \times n\) where \(d\) is the dimension of the final layer and \(n\)
is the number of classes. After using only \emph{influential features}
it drops significantly. For example for Resnet20 and CIFAR-10, the
computational complexity for the final layer is \(64 \times 10\) which
drops to \(5 \times 10\) with \emph{influential features}.

\vspace{-3em}

\hypertarget{decomposed-softmax}{%
\subsubsection{Decomposed Softmax}\label{decomposed-softmax}}

In a standard CNN, for a given data input \(x \in \mathcal{X}\), the
prediction is given as \(\sigma(WI^d)\) where \(I^d\) is the input at
depth \(d\) and \(W \in \mathbbm{R}^{m \times n}\). For the input, the
output is a floating point representation normalized by the softmax. The
probability for each \(y_i\) is then:

\(P(y_i| x) \sim \sum_j W_{ij} x_j \equiv <W_i x>\), and the predicted
class is:

\(P(\hat{y} = y_i| x) \sim \underset{i}{\operatorname{argmax}} < W_i x >\)

For predicting with influential features the probabilities are computed
as:

\(P(\hat{y} = y_i| x) \sim \underset{i}{\operatorname{argmax}} < \mathbf{w}_{inf} x >\)
where \(\mathbf{w}_{inf} \in \mathbbm{R}^{k}, \ k << m\)

The output thus can be thought of as a \emph{decomposed softmax}. An
illustration of this is given in Fig \ref{fig:pred_inf}

\hypertarget{visualizing-influential-features}{%
\subsubsection{Visualizing Influential
Features}\label{visualizing-influential-features}}

Here we visualize the features from retrained model Resnet50 after
decomposition and compare it with the standard Resnet50 trained on
Imagenet.

For each image we subtract the mean of the influential filters at the
final convolutional layer after average pooling at each dimension and
upsample the activation to \(224 \times 224\) which is the standard size
of the input image to Resnet models. For comparison we select \(k\)
filters from other filters for the same class The results are in Table
\ref{tab:vis_compare_feats}, where standard Resnet50 features trained on
Imagenet are displayed alongside \emph{influential features} and
\emph{non-influential features} on the decomposed Resnet50.

\begin{table}[!h]
  \centering
  \vspace{-1in}
  \begin{tabular}{ccccc}
    \hline
    \bfseries Category & \bfseries Image & \bfseries Standard Resnet &\bfseries Non-Influential & \bfseries Influential\\
                       & &\bfseries Features &\bfseries Features & \bfseries Features \\
    \hline
    \bfseries Abacus & \includegraphics[height=1.4cm,width=2cm]{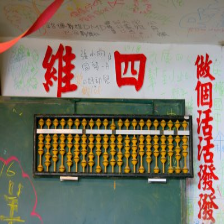} & \includegraphics[height=1.4cm,width=2cm]{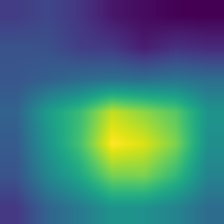} & \includegraphics[height=1.4cm,width=2cm]{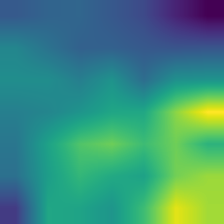} & \includegraphics[height=1.4cm,width=2cm]{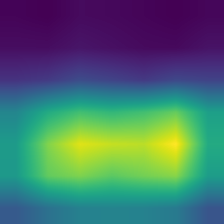} \\~\\
    \hline
    \bfseries Seat belt & \includegraphics[height=1.4cm,width=2cm]{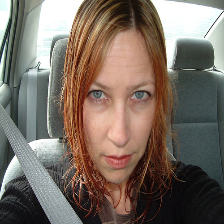} & \includegraphics[height=1.4cm,width=2cm]{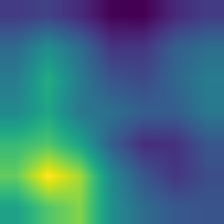} & \includegraphics[height=1.4cm,width=2cm]{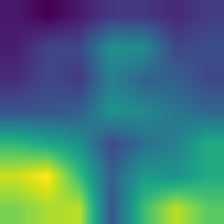} & \includegraphics[height=1.4cm,width=2cm]{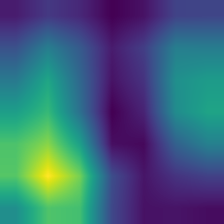} \\~\\
    \hline
    \bfseries Squirrel monkey & \includegraphics[height=1.4cm,width=2cm]{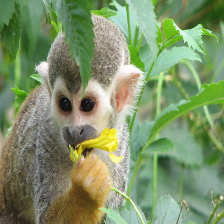} & \includegraphics[height=1.4cm,width=2cm]{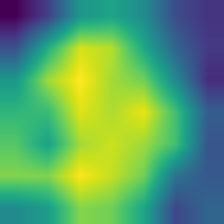} & \includegraphics[height=1.4cm,width=2cm]{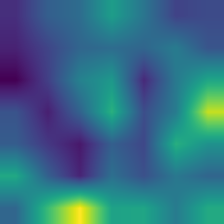} & \includegraphics[height=1.4cm,width=2cm]{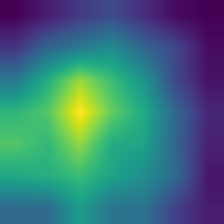} \\~\\
    \hline
    \bfseries Wreck & \includegraphics[height=1.4cm,width=2cm]{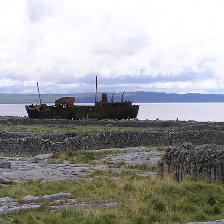} & \includegraphics[height=1.4cm,width=2cm]{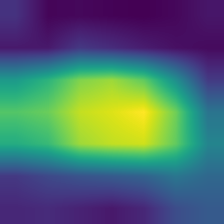} & \includegraphics[height=1.4cm,width=2cm]{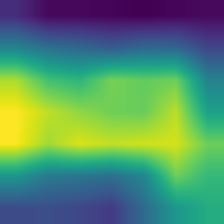} & \includegraphics[height=1.4cm,width=2cm]{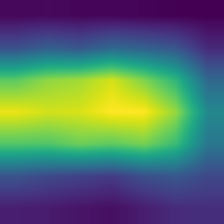} \\~\\
    \hline
    \bfseries Madagascar cat & \includegraphics[height=1.4cm,width=2cm]{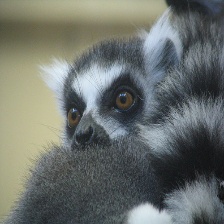} & \includegraphics[height=1.4cm,width=2cm]{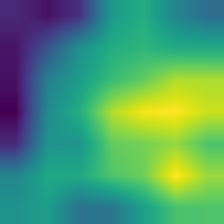} & \includegraphics[height=1.4cm,width=2cm]{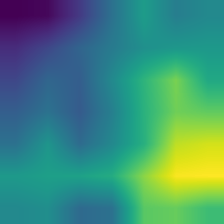} & \includegraphics[height=1.4cm,width=2cm]{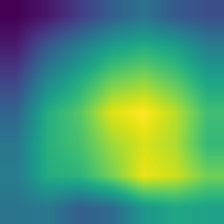} \\~\\
    \hline
    \bfseries Standard poodle & \includegraphics[height=1.4cm,width=2cm]{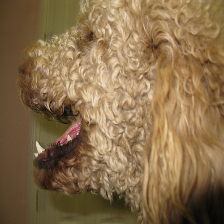} & \includegraphics[height=1.4cm,width=2cm]{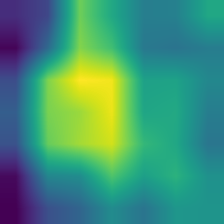} & \includegraphics[height=1.4cm,width=2cm]{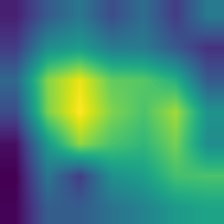} & \includegraphics[height=1.4cm,width=2cm]{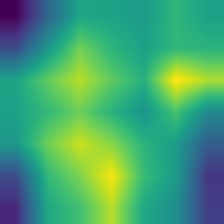} \\~\\
    \hline
    \bfseries Welsh springer spaniel & \includegraphics[height=1.4cm,width=2cm]{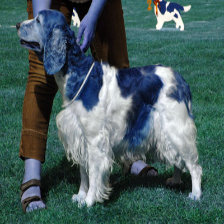} & \includegraphics[height=1.4cm,width=2cm]{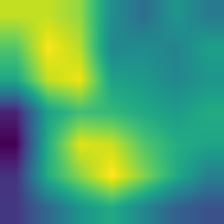} & \includegraphics[height=1.4cm,width=2cm]{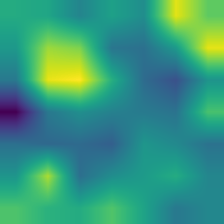} & \includegraphics[height=1.4cm,width=2cm]{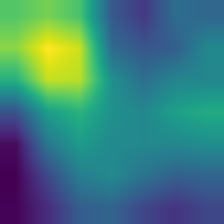} \\~\\
    \hline
    \bfseries Frilled lizard & \includegraphics[height=1.4cm,width=2cm]{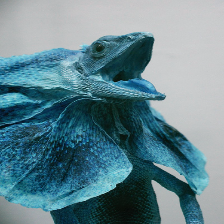} & \includegraphics[height=1.4cm,width=2cm]{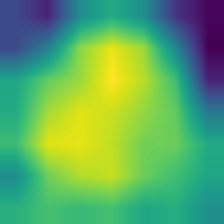} & \includegraphics[height=1.4cm,width=2cm]{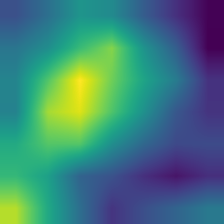} & \includegraphics[height=1.4cm,width=2cm]{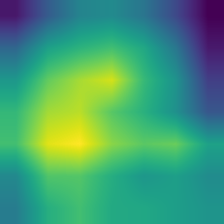} \\~\\
    \hline
    \bfseries Thatch & \includegraphics[height=1.4cm,width=2cm]{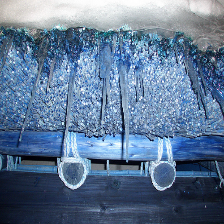} & \includegraphics[height=1.4cm,width=2cm]{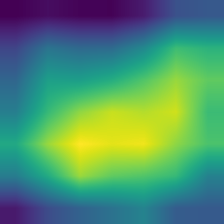} & \includegraphics[height=1.4cm,width=2cm]{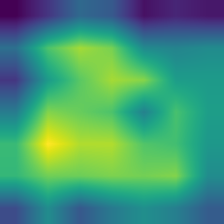} & \includegraphics[height=1.4cm,width=2cm]{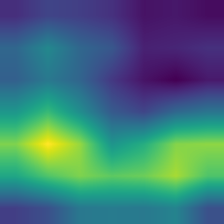} \\~\\
    \hline
  \end{tabular}
  \caption{Comparison of visualization of standard Resnet and \emph{influential}
    and \emph{non-influential} features}
  \label{tab:vis_compare_feats}
\end{table}

We can see that the \emph{influential features} have better overlap with
the regions for the target class of the objects in the scene while the
\emph{non-influential} features seem to focus more on the environment.
This lends our hypothesis of these features being critical to the object
classification.

\hypertarget{essentiality-of-influential-features}{%
\subsubsection{Essentiality of Influential
Features}\label{essentiality-of-influential-features}}

Here we perform an ablation study with and without the \emph{influential
features} to demonstrate their essential nature. For the study we
replace the values at pre-final layer of \emph{influential features} and
\emph{non-influential features} respectively with Gaussian Noise.
\cite{liang2020training} zero out the values of the corresponding
features but that would create a more drastic perturbation which is sure
to affect the result negatively. We feel introducing Gaussian Noise is
instead a better way to check the features' essential nature. The
results are in Table \ref{tab:comp_noise_inf}.

\begin{table}
  \centering
  \begin{tabular}{llll}
    \hline
    \textbf{Model} & \textbf{Original} & \textbf{Accuracy with} & \textbf{Accuracy with}\\
                   & \textbf{Accuracy} & \textbf{Influential Features} & \textbf{Non-Influential Features}\\
                   & & \textbf{replaced with} & \textbf{replaced with}\\
                   & & \textbf{Gaussian Noise} & \textbf{Gaussian Noise}\\
    \hline
    Resnet20 & 91.17 & 48.97 & 82.24\\
    \hline
    Resnet50 & 74.55 & 12.52 & 74.40\\
    \hline
  \end{tabular}
  \caption{Comparison with injecting noise in Influential Features vs Non-Influential Features}
  \label{tab:comp_noise_inf}
\end{table}

For Resnet20 and CIFAR-10 the accuracy by replacing \emph{influential
features} with noise drops from
\(91.17%\) to \(48.97%\), a drop of \(42.2%
\), while replacing \emph{non-influential features} results in an
accuracy drop of \(8.97%
\). For Resnet50 and Imagenet, the results are even more pronounced with
accuracy drop with noise for \emph{influential features} resulting in a
drop of \(52.05%\) and only a \(0.15%
\) drop for corresponding \emph{non-influential features}.

This shows that \emph{influential features} are essential to the correct
classification for the given class. We can explain the difference in
results between CIFAR-10 and Imagenet due to lower resolution of
CIFAR-10 images, where object features in the images themselves are not
so pronounced.

\hypertarget{discussion-and-future-work}{%
\section{Discussion and Future Work}\label{discussion-and-future-work}}

We have described here a novel method of identifying the \(k\) most
influential features for the final layers of a CNN. The method allows us
greater interpretability into the CNN and also to decompose the final
layer into separate class specific final layer. We have shown that these
features are essential for classifcation for the given labels. We have
also shown that deeper networks have inherently more disentangled
representations in the final layers. Finally, we also provide the
implementation for the experiments and the decomposed final layer
computations.

For future work, the focus of the investigation can shift to inner
layers of the CNN. We have also seen that the accuracy can also be
recovered when trained for very few epochs for CIFAR-10. This raises the
possibility of further hierarchical decomposition of the CNN as a tree
like structure which can be explored in future work.

Another aspect that can be explored in the future is the overlap between
influential features and the object itself in the image. These features
may be used for Semantic Segmentation of objects without training
segmentation annotations. That might require a reformulation of the task
with additional constraints added to classification.

Models like Efficientnet \cite{tan2019efficientnet} give us different
results as they are derived from Neural Architecture Search (NAS) which
leads to more complex models and we can hypothesize that their
representations are more entangled. Other models on the other hand are
designed by human intuition and are simpler. However, that would require
exploration of NAS based models which we also leave to future work.

\vspace{-1em}

\hypertarget{acknowledgements}{%
\section{Acknowledgements}\label{acknowledgements}}

We wish to thank the MEITY, Govt. of India, which funded part of this
research with Visvesvaraya PhD Fellowship award number MEITY-PHD-1035.

\vspace{-1em}

%% Bibliography, only natbib for elsevier
\bibliography{/home/joe/Documents/pandoc/decomposing-the-deep-finding-class-specific-filters-in-deep-cnns-v1/decomposing-the-deep-5840ae846e.bib}

\begin{thebibliography}{10}
\expandafter\ifx\csname url\endcsname\relax
  \def\url#1{\texttt{#1}}\fi
\expandafter\ifx\csname urlprefix\endcsname\relax\def\urlprefix{URL }\fi
\expandafter\ifx\csname href\endcsname\relax
  \def\href#1#2{#2} \def\path#1{#1}\fi

\bibitem{szegedy2015going}
C.~Szegedy, W.~Liu, Y.~Jia, P.~Sermanet, S.~Reed, D.~Anguelov, D.~Erhan,
  V.~Vanhoucke, A.~Rabinovich, {Going} {Deeper} {with} {Convolutions}, in:
  {Proc.} {of} {the} {IEEE} {Conf.} {on} {Comput.} {Vis.} {and} {Pattern}
  {Recognit.}, 2015, pp. 1--9.

\bibitem{he2016deep}
K.~He, X.~Zhang, S.~Ren, J.~Sun, {Deep} {Residual} {Learning} {for} {Image}
  {Recognition}, in: {Proc.} {of} {the} {IEEE} {Conf.} {on} {Comput.} {Vis.}
  {and} {Pattern} {Recognit.}, 2016, pp. 770--778.

\bibitem{radford2019language}
A.~Radford, J.~Wu, R.~Child, D.~Luan, D.~Amodei, I.~Sutskever, {Language}
  {Models} {are} {Unsupervised} {Multitask} {Learners},
  \url{https://cdn.openai.com/better-language-models/language_models_are_unsupervised_multitask_learners.pdf},
  accessed: 2021-06-25 (2019).

\bibitem{brown2020language}
T.~B. Brown, B.~Mann, N.~Ryder, M.~Subbiah, J.~Kaplan, P.~Dhariwal,
  A.~Neelakantan, P.~Shyam, G.~Sastry, A.~Askell, S.~Agarwal, A.~Herbert-Voss,
  G.~Krueger, T.~Henighan, R.~Child, A.~Ramesh, D.~M. Ziegler, J.~Wu,
  C.~Winter, C.~Hesse, M.~Chen, E.~Sigler, M.~Litwin, S.~Gray, B.~Chess,
  J.~Clark, C.~Berner, S.~McCandlish, A.~Radford, I.~Sutskever, D.~Amodei,
  {Language} {Models} {Are} {Few} {-} {Shot} {Learners}, in: {Adv.} {in}
  {Neural} {Inf.} {Process.} {Syst.}, 2020.

\bibitem{choromaska2015loss}
A.~Choromańska, M.~Henaff, M.~Mathieu, G.~B. Arous, Y.~LeCun, {The} {Loss}
  {Surfaces} {of} {Multilayer} {Networks}, in: {Proc.} {of} {the}
  {International} {Conf.} {on} {Artif.} {Intell.} {and} {Stat.}, Vol.~38, 2015,
  pp. 192--204.

\bibitem{du2019gradient}
S.~Du, X.~Zhai, B.~Póczos, A.~Singh, {Gradient} {Descent} {Provably}
  {Optimizes} {over} {-} {Parameterized} {Neural} {Networks}, in: {Proc.} {of}
  {the} {International} {Conf.} {on} {Learn.} {Represent.}, 2019.

\bibitem{du2019gradient_a}
S.~Du, J.~Lee, H.~Li, L.~Wang, X.~Zhai, {Gradient} {Descent} {Finds} {Global}
  {Minima} {of} {Deep} {Neural} {Networks}, in: {Proc.} {of} {the}
  {International} {Conf.} {on} {Mach.} {Learn.}, 2019.

\bibitem{huang2017densely}
G.~Huang, Z.~Liu, K.~Q. Weinberger, {Densely} {Connected} {Convolutional}
  {Networks}, in: {Proc.} {of} {the} {IEEE} {Conf.} {on} {Comput.} {Vis.} {and}
  {Pattern} {Recognit.}, 2017, pp. 2261--2269.

\bibitem{tan2019efficientnet}
M.~Tan, Q.~V. Le, {Efficientnet:} {Rethinking} {Model} {Scaling} {for}
  {Convolutional} {Neural} {Networks}, in: {Proc.} {of} {the} {International}
  {Conf.} {on} {Mach.} {Learn.}, 2019.

\bibitem{zhang2020survey_a}
Y.~Zhang, P.~Tiño, A.~Leonardis, K.~Tang, {A} {Survey} {on} {Neural} {Network}
  {Interpretability}, {IEEE} {Trans.} {on} {Emerging} {Topics} {in}
  {Computational} {Intell.} (2020).

\bibitem{frosst2017distilling}
N.~Frosst, G.~E. Hinton, {Distilling} {a} {Neural} {Network} {into} {a} {Soft}
  {Decision} {Tree}, in: {Cex@ai*ia}, 2017.

\bibitem{zeiler2014visualizing}
M.~D. Zeiler, R.~Fergus, {Visualizing} {and} {Understanding} {Convolutional}
  {Networks}, in: {Proc.} {of} {the} {Eur.} {Conf.} {on} {Comput.} {Vis.},
  2014.

\bibitem{szegedy2014intriguing}
C.~Szegedy, W.~Zaremba, I.~Sutskever, J.~Bruna, D.~Erhan, I.~Goodfellow,
  R.~Fergus, {Intriguing} {Properties} {of} {Neural} {Networks}, in: {Proc.}
  {of} {the} {International} {Conf.} {on} {Learn.} {Represent.}, 2014.

\bibitem{simonyan2014deep}
K.~Simonyan, A.~Vedaldi, A.~Zisserman, {Deep} {Inside} {Convolutional}
  {Networks:} {Visualising} {Image} {Classification} {Models} {and} {Saliency}
  {Maps}, in: {Proc.} {of} {the} {International} {Conf.} {on} {Learn.}
  {Represent.}, 2014.

\bibitem{gonzalezgarcia2017semantic}
A.~Gonzalez-Garcia, D.~Modolo, V.~Ferrari, {Do} {Semantic} {Parts} {Emerge}
  {in} {Convolutional} {Neural} {Networks?}, {International} {J.} {of}
  {Comput.} {Vis.} 126 (2017) 476--494.

\bibitem{bau2017network}
D.~Bau, B.~Zhou, A.~Khosla, A.~Oliva, A.~Torralba, {Network} {Dissection} {-}
  {Quantifying} {Interpretability} {of} {Deep} {Visual} {Representations.}, in:
  {Proc.} {of} {the} {IEEE} {Conf.} {on} {Comput.} {Vis.} {and} {Pattern}
  {Recognit.}, 2017, pp. 3319--3327.

\bibitem{bach2015pixelwise}
S.~Bach, A.~Binder, G.~Montavon, F.~Klauschen, K.~Müller, W.~Samek, {On}
  {Pixel} {-} {Wise} {Explanations} {for} {Non} {-} {Linear} {Classifier}
  {Decisions} {by} {Layer} {-} {Wise} {Relevance} {Propagation}, in: {Plos}
  {One}, 2015.

\bibitem{hendricks2016generating}
L.~A. Hendricks, Z.~Akata, M.~Rohrbach, J.~Donahue, B.~Schiele, T.~Darrell,
  {Generating} {Visual} {Explanations}, in: {Proc.} {of} {the} {Eur.} {Conf.}
  {on} {Comput.} {Vis.}, 2016.

\bibitem{alain2017understanding}
G.~Alain, Y.~Bengio, {Understanding} {Intermediate} {Layers} {Using} {Linear}
  {Classifier} {Probes}, in: {Proc.} {of} {the} {International} {Conf.} {on}
  {Learn.} {Represent.}, 2017.

\bibitem{zhang2018interpretable}
Q.~Zhang, Y.~Wu, S.-C. Zhu, {Interpretable} {Convolutional} {Neural}
  {Networks}, in: {Proc.} {of} {the} {IEEE} {Conf.} {on} {Comput.} {Vis.} {and}
  {Pattern} {Recognit.}, 2018, pp. 8827--8836.

\bibitem{liang2020training}
H.~Liang, Z.~Ouyang, Y.~Zeng, H.~Su, Z.~He, S.-T. Xia, J.~Zhu, B.~Zhang,
  {Training} {Interpretable} {Convolutional} {Neural} {Networks} {by}
  {Differentiating} {Class} {-} {Specific} {Filters}, in: {Proc.} {of} {the}
  {Eur.} {Conf.} {on} {Comput.} {Vis.}, Springer International Publishing,
  2020, pp. 622--638.

\bibitem{yang2019quantization}
J.~Yang, X.~Shen, J.~Xing, X.~Tian, H.~Li, B.~Deng, J.~Huang, X.~Hua,
  {Quantization} {Networks}, in: {Proc.} {of} {the} {IEEE} {Conf.} {on}
  {Comput.} {Vis.} {and} {Pattern} {Recognit.}, 2019.

\bibitem{zhang2018interpreting}
Q.~Zhang, R.~Cao, F.~Shi, Y.~Wu, S.~Zhu, {Interpreting} {CNN} {Knowledge} {via}
  {an} {Explanatory} {Graph}, in: {Proc.} {of} {the} {AAAI} {Conf.} {on}
  {Artif.} {Intell.}, 2018.

\bibitem{zhang2019interpreting}
Q.~Zhang, Y.~Yang, Y.~Wu, S.-C. Zhu, {Interpreting} {CNNs} {via} {Decision}
  {Trees}, in: {Proc.} {of} {the} {IEEE} {Conf.} {on} {Comput.} {Vis.} {and}
  {Pattern} {Recognit.}, 2019, pp. 6254--6263.

\bibitem{zhou2016learning}
B.~Zhou, A.~Khosla, A.~Lapedriza, A.~Oliva, A.~Torralba, {Learning} {Deep}
  {Features} {for} {Discriminative} {Localization}, in: {Proc.} {of} {the}
  {IEEE} {Conf.} {on} {Comput.} {Vis.} {and} {Pattern} {Recognit.}, IEEE, 2016.

\bibitem{selvaraju2017gradcam}
R.~R. Selvaraju, M.~Cogswell, A.~Das, R.~Vedantam, D.~Parikh, D.~Batra, {Grad}
  {-} {CAM:} {Visual} {Explanations} {from} {Deep} {Networks} {via} {Gradient}
  {-} {Based} {Localization}, in: {Proc.} {of} {the} {IEEE} {International}
  {Conf.} {on} {Comput.} {Vis.}, 2017.

\bibitem{wickramanayake2019flex}
S.~Wickramanayake, W.~Hsu, M.~Lee, {FLEX:} {Faithful} {Linguistic}
  {Explanations} {for} {Neural} {Net} {Based} {Model} {Decisions}, in: {Proc.}
  {of} {the} {AAAI} {Conf.} {on} {Artif.} {Intell.}, 2019, pp. 2539--2546.

\bibitem{wickramanayake2021comprehensible}
S.~Wickramanayake, W.~Hsu, M.~Lee, {Comprehensible} {Convolutional} {Neural}
  {Networks} {via} {Guided} {Concept} {Learning}, in: {IJCNN}, 2021.

\bibitem{liu2015sparse}
B.~Liu, M.~Wang, H.~Foroosh, M.~Tappen, M.~Pensky, {Sparse} {Convolutional}
  {Neural} {Networks}, in: {Proc.} {of} {the} {IEEE} {Conf.} {on} {Comput.}
  {Vis.} {and} {Pattern} {Recognit.}, 2015, pp. 806--814.

\bibitem{li2017pruning_a}
H.~Li, A.~Kadav, I.~Durdanovic, H.~Samet, H.~Graf, {Pruning} {Filters} {for}
  {Efficient} {ConvNets}, in: {Proc.} {of} {the} {International} {Conf.} {on}
  {Learn.} {Represent.}, 2017.

\bibitem{kumar2021pruning}
A.~Kumar, A.~M. Shaikh, Y.~Li, H.~Bilal, B.~Yin, {Pruning} {Filters} {with} {L1
  {-} {Norm}} {and} {Capped} {L1 {-} {Norm}} {for} {CNN} {Compression},
  {Applied} {Intell.} 51 (2021) 1152--1160.

\bibitem{lin2020toward}
S.~Lin, R.~Ji, Y.~Li, C.~Deng, X.~Li, {Toward} {Compact} {ConvNets} {via}
  {Structure} {-} {Sparsity} {Regularized} {Filter} {Pruning}, in: {IEEE}
  {Trans.} {on} {Neural} {Networks} {and} {Learn.} {Syst.}, 2020.

\bibitem{li2019exploiting}
Y.~Li, S.~Lin, B.~Zhang, J.~Liu, D.~Doermann, Y.~Wu, F.~Huang, R.~Ji,
  {Exploiting} {Kernel} {Sparsity} {and} {Entropy} {for} {Interpretable} {CNN}
  {Compression}, in: {Proc.} {of} {the} {IEEE} {Conf.} {on} {Comput.} {Vis.}
  {and} {Pattern} {Recognit.}, 2019.

\bibitem{wang2018learning}
Y.~Wang, V.~I. Morariu, L.~Davis, {Learning} {a} {Discriminative} {Filter}
  {Bank} {Within} {a} {CNN} {for} {Fine} {-} {Grained} {Recognition}, in:
  {Proc.} {of} {the} {IEEE} {Conf.} {on} {Comput.} {Vis.} {and} {Pattern}
  {Recognit.}, 2018, pp. 4148--4157.

\bibitem{jiang2017learning}
Z.~Jiang, Y.~Wang, L.~Davis, W.~Andrews, V.~Rozgic, {Learning} {Discriminative}
  {Features} {via} {Label} {Consistent} {Neural} {Network}, in: {Proc.} {of}
  {the} {IEEE} {Winter} {Conf.} {on} {Appl.} {of} {Comput.} {Vis.}, 2017, pp.
  207--216.

\bibitem{ioffe2015batch}
S.~Ioffe, C.~Szegedy, {Batch} {Normalization:} {Accelerating} {Deep} {Network}
  {Training} {by} {Reducing} {Internal} {Covariate} {Shift}, in: {Proc.} {of}
  {the} {International} {Conf.} {on} {Mach.} {Learn.}, 2015.

\bibitem{krizhevsky2012imagenet}
A.~Krizhevsky, I.~Sutskever, G.~E. Hinton, {Imagenet} {Classification} {with}
  {Deep} {Convolutional} {Neural} {Networks}, in: {Adv.} {in} {Neural} {Inf.}
  {Process.} {Syst.}, 2012, pp. 1097--1105.

\bibitem{krizhevsky2009learning}
A.~Krizhevsky, {Learning} {Multiple} {Layers} {of} {Features} {from} {Tiny}
  {Images},
  \url{https://www.cs.toronto.edu/~kriz/learning-features-2009-TR.pdf} (2009).

\bibitem{deng2009imagenet}
J.~Deng, W.~Dong, R.~Socher, L.-J. Li, K.~Li, L.~Fei-Fei, {Imagenet:} {A}
  {Large} {-} {Scale} {Hierarchical} {Image} {Database}, in: {Proc.} {of} {the}
  {IEEE} {Conf.} {on} {Comput.} {Vis.} {and} {Pattern} {Recognit.}, 2009, pp.
  248--255.

\bibitem{zoph2017neural}
B.~Zoph, Q.~V. Le, {Neural} {Architecture} {Search} {with} {Reinforcement}
  {Learning}, in: {Proc.} {of} {the} {International} {Conf.} {on} {Learn.}
  {Represent.}, 2017.

\bibitem{zagoruyko2016wide}
S.~Zagoruyko, N.~Komodakis, {Wide} {Residual} {Networks}, in: {Proc.} {of}
  {the} {Br.} {Mach.} {Vis.} {Conf.}, 2016.

\end{thebibliography}

%% Post includes
\end{document}